\documentclass{article}

\usepackage{PRIMEarxiv}
\usepackage[utf8]{inputenc} 
\usepackage[T1]{fontenc}    
\usepackage{hyperref}       
\usepackage{url}            
\usepackage{booktabs}       
\usepackage{amsmath,amssymb,amsfonts}       
\usepackage{nicefrac}       
\usepackage{microtype}      
\usepackage{lipsum}
\usepackage{fancyhdr}       
\usepackage{graphicx}       
\graphicspath{{media/}}     
\usepackage{multirow}
\usepackage{subcaption}
\usepackage{threeparttable}

\title{Zero-shot information extraction from radiological reports using ChatGPT
\thanks{\textit{\underline{Citation}}: 
\textbf{Authors. Title. Pages.... DOI:000000/11111.}} 
}

\author{
  Danqing Hu \\
  Zhejiang Lab \\
  Hangzhou, Zhejiang, China\\
  \texttt{hudq@zhejianglab.com} \\
   \And
  Bing Liu\\
  Peking University Cancer Hospital and Institute \\
  Beijing, China\\
  \texttt{liubing983811735@126.com} \\  
   \And
  Xiaofeng Zhu \\
  Zhejiang Lab \\
  Hangzhou, Zhejiang, China\\
  \texttt{andy.zhu@zhejianglab.com} \\
   \And
  Xudong Lu \\
  Zhejiang University \\
  Hangzhou, Zhejiang, China\\
  \texttt{lvxd@zju.edu.cn} \\
  \And
  Nan Wu \\
  Peking University Cancer Hospital and Institute \\
  Beijing, China\\
  \texttt{nanwu@bjmu.edu.cn} \\
}

\begin{document}
\maketitle

\begin{abstract}

Electronic health records contain an enormous amount of valuable information, but many are recorded in free text. Information extraction is the strategy to transform the sequence of characters into structured data, which can be employed for secondary analysis. However, the traditional information extraction components, such as named entity recognition and relation extraction, require annotated data to optimize the model parameters, which has become one of the major bottlenecks in building information extraction systems. With the large language models achieving good performances on various downstream NLP tasks without parameter tuning, it becomes possible to use large language models for zero-shot information extraction. In this study, we aim to explore whether the most popular large language model, ChatGPT, can extract useful information from the radiological reports. We first design the prompt template for the interested information in the CT reports. Then, we generate the prompts by combining the prompt template with the CT reports as the inputs of ChatGPT to obtain the responses. A post-processing module is developed to transform the responses into structured extraction results. We conducted the experiments with 847 CT reports collected from Peking University Cancer Hospital. The experimental results indicate that ChatGPT can achieve competitive performances for some extraction tasks compared with the baseline information extraction system, but some limitations need to be further improved.

\end{abstract}


\section{Introduction}

With the rapid development of electronic health records (EHRs), a large amount of medical data has been accumulated, which contains valuable information for disease diagnosis and prognosis prediction \cite{Xiao2018,Hu2023,Hu2020}. However, free-text data, such as radiological reports, pathological reports, and operation notes, cannot be directly used by many algorithms, which seriously hinders the secondary utilization of medical data \cite{Yadav2018,Yim2016,Shickel2018}. Manually extracting structured information from free-text data is time-consuming, labor-intensive, error-prone, and expensive. Therefore, information extraction (IE) came into being to solve this problem by automatically extracting structured information from unstructured text \cite{Wang2018,Datta2019}.

IE is commonly regarded as a specialized field of natural language processing (NLP), referring to automatic extraction of entities, concepts, and events as well as the relations between them from free text \cite{Wang2018}. An IE system usually consists of one or more of the following subtasks: named entity recognition (NER) that identifies entity names from text \cite{Lei2013,Li2022}, coreference resolution that finds all expressions referring to the same entity in text \cite{Liu2023}, relation extraction that associates entities with different relations \cite{Nasar2021}, post-processing that transforms the extracted entities and relations into structured information. To develop the IE system, two main categories of methods were employed, i.e., rule-based and machine learning.

Rule-based IE systems typically include rules and an interpreter to run the rules \cite{Wang2018}. Regular expression is the most common form of the rule, which can find the particular pattern in the text and extract the keywords in this pattern. A rule-based IE system is usually composed of many rules to achieve a high recall value because of the diverse expressions of the same semantics in natural language \cite{Friedman1994,Savova2010,Aronson2010}. However, these rules may capture information with the same pattern but different semantics, thus leading to low precision value \cite{Saeed2016}. Besides, the rules are typically written by engineers with domain expert knowledge or clinicians, which is very challenging to maintain, update, and expand as rules become more numerous and complex \cite{Saeed2016}.

Machine learning-based IE approaches alleviate the dilemma of manually developing rules by automatically learning semantic patterns from data \cite{Wang2018}. These methods typically use the hand-crafted features from the free text as the input and combine with machine learning algorithms such as conditional random fields, logistic regression, and support vector machine to identify the named entities and their relations \cite{Saeed2016,Roberts2012,Abeed2015,Li2015}. As deep learning demonstrates powerful representation learning capabilities, word embedding methods like word2vec and glove have gradually replaced the hand-crafted features and combined with recurrent neural networks to achieve superior performances in various NLP tasks \cite{Unanue2017,Gao2017,Hu2021}. With the improvement of computing power and the proposal of the self-attention mechanism, the pre-trained large models like BERT further improve the performances through pre-training using the self-supervised strategy on massive text data and then fine-tuning on task-specific labeled data \cite{Xiaohui2019,Zhang2021,Surabhi2022,Chen2023}. Although the machine learning approaches show good information extraction results, these methods still require enormous annotated data, which is hectic, labor-intensive, and time-consuming.

Recent works on pre-trained large language models (LLM), such as GPT-3 and ChatGPT, suggest that LLMs perform well on various downstream NLP tasks even without parameter tuning \cite{Brown2020,Min2023}. The LLMs receive the prompts with certain content and instructions and then provide the responses in a question-answering manner, which may become a new IE paradigm \cite{Agrawal2022,Wei2023,Hu20232}. In this study, we aim to explore whether the ChatGPT can do zero-shot IE from radiological reports. We first design the prompt template for the IE task. Then, we use the prompt template and radiological reports to generate the prompts as the inputs for ChatGPT to obtain the responses. After that, we develop a post-processing module to transform the extracted results into a structured form. We conduct the experiments using real clinical radiological reports collected from Peking University Cancer Hospital. Experimental results show that ChatGPT can achieve competitive performance in extracting radiological report information but still have some limitations to be further improved.

\section{Materials and methods}
\label{sec:Methods}

\subsection{Data}

We collected a total of 847 computed tomography (CT) reports of lung cancer patients treated from 2010 to 2018 at the Department of Thoracic Surgery II, Peking University Cancer Hospital. All patients underwent a chest CT scan within two months before curative resections. And we collected these preoperative CT reports as our clinical text data. Under the clinician's instructions, we defined 11 lung cancer-relation questions for the IE task, and all questions are shown in Table \ref{table1}. We recruited two engineers with a medical informatics background to label the answers manually. A clinician reviewed the different annotated labels to determine the final results as the gold standard. We obtained approval from the ethics committee of Peking University Cancer Hospital (2022KT128) before conducting this study.

\begin{table}[t]
\caption{Lung cancer-related questions for information extraction.}
\label{table1}
\begin{center}
\begin{tabular}{cll}
\toprule
    No.     & Question                                    & Answer type \\
\midrule
    1       & Tumor location                              & Categorical \\
    2       & Tumor long diameter                         & Numerical   \\
    3       & Tumor short diameter                        & Numerical   \\
    4       & Is the tumor solid                          & Boolean     \\
    5       & Is the tumor ground-glass opacity           & Boolean     \\
    6       & Is the tumor mixed ground-glass opacity     & Boolean     \\
    7       & Does the tumor have spiculations            & Boolean     \\
    8       & Does the tumor have lobulations             & Boolean     \\
    9       & Is there pleural invasion or indentation    & Boolean     \\
    10      & Are mediastinal lymph nodes enlarged        & Boolean     \\
    11      & Are hilar lymph nodes enlarged              & Boolean     \\
\bottomrule
\end{tabular}
\end{center}

\end{table}

\begin{figure}[h]
\centering
\includegraphics[width=\textwidth]{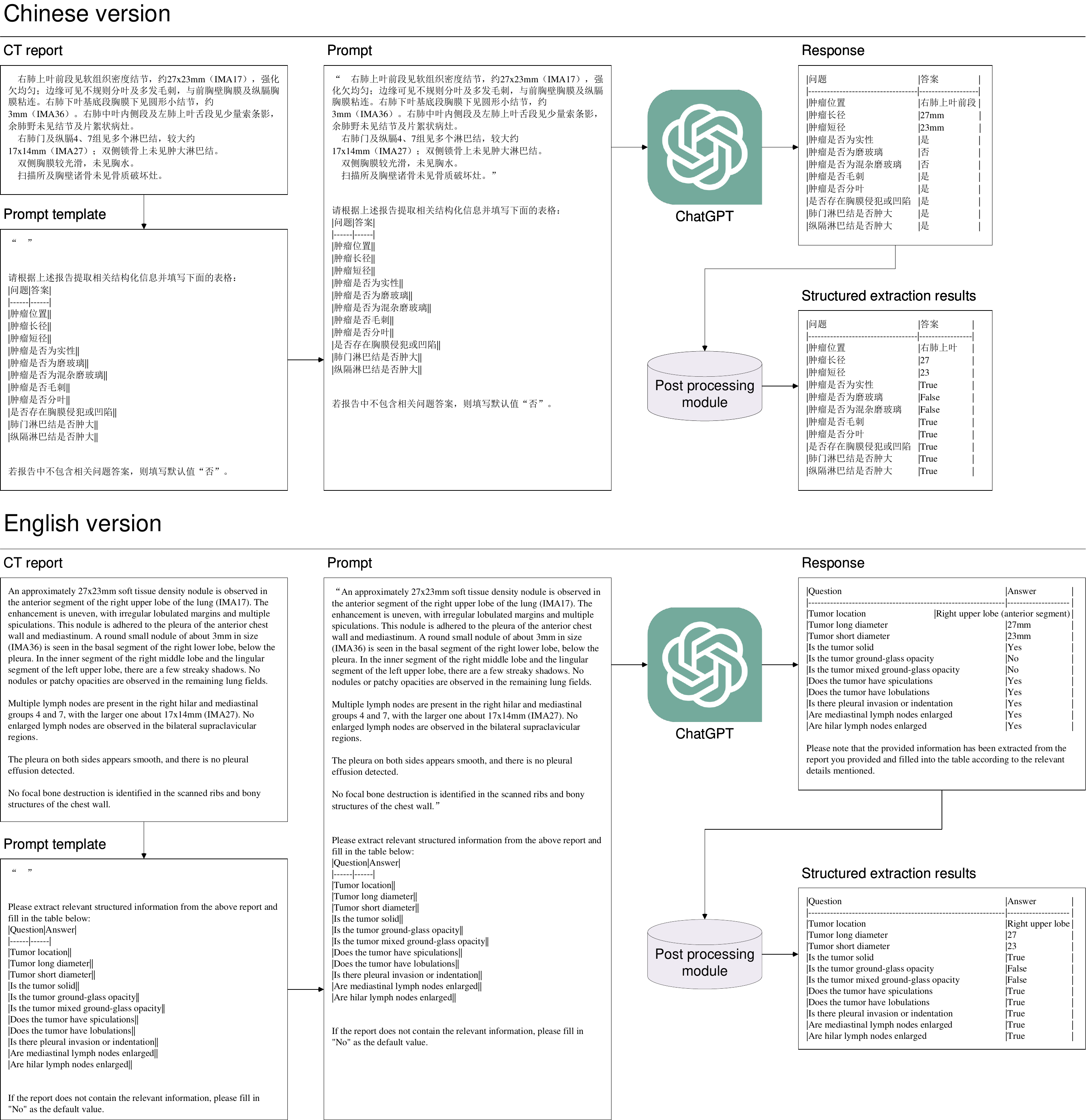}
\caption{Question answering framework for zero-shot IE.}
\label{figure1}
\end{figure}

\subsection{Question answering framework for zero-shot IE}

\subsubsection{Prompt template for zero-shot IE}

Based on the questions in Table \ref{table1}, we first design the IE prompt template shown in Figure \ref{figure1}. The prompt template is composed of three parts. The first part is the original CT report. The second part consists of an IE instruction and an unfilled question form. In this study, we design the question form to ask ChatGPT to fill in the blanks based on the provided CT report. The filled question form is convenient for locating and post-processing extraction results. The third part is some supplementary requirements for the IE task. In this study, we ask ChatGPT to set the answer to False by default for questions not mentioned in the CT report. In the prompt template, we only provide the CT report to be extracted and do not provide any sample CT reports and extracted results to extract information in a zero-shot manner.

\subsubsection{Question answering using ChatGPT}

Using the prompt template, we combine the CT reports to generate the prompts for question answering. We input the prompt to the ChatGPT via the web page and obtain its response. For each CT report, we restart a new dialogue to avoid the impact of extraction results in the previous dialogue. As a language model, the response of ChatGPT not only contains the filled question form we want but also some other content we are not interested in. So, we only extract the content in the filled question form as the outputs. The question-answering procedure is shown in Figure \ref{figure1}.

\subsubsection{Post-processing for structured IE}

Note that although we only extract the content in the filled question form as the outputs of IE, the answers are not always structured because a language model may generate various sequences. To solve this problem, we develop a rule-based post-processing module with regular expressions to transform the unstructured answers into structured ones. Specifically, we utilize keywords such as "right", "left", "upper", "middle", and "lower" in the extracted tumor location to convert the extracted results to 6 formatted answers, i.e., right upper lobe, right middle lobe, right lower lobe, left upper lobe, left lower lobe, and others. For tumor long and short diameters, we first extract the float number and the length unit and then unify the float lengths to be in millimeters. For three tumor density-related questions, we determine their boolean values based on the predefined priority to guarantee that they are mutually exclusive. For remaining questions, we utilize the keywords "yes" and "No" in the answers to determine the boolean values. Note that ChatGPT often answers "Not mentioned" when it finds nothing relevant in the CT report, even though we have provided the instruction in part III of the prompt template. So, we set the boolean value to False when "Not mentioned" in the answer. Figure \ref{figure1} shows the post-processing procedure in the overall question-answering framework for zero-shot IE.

\subsection{Experimental Setup}

In this study, we manually designed the prompt template and generated the prompts for IE. All the question-answering dialogues were finished via the official ChatGPT web (https://chat.openai.com). We collected the filled question forms in the responses of ChatGPT and employed the post-processing module to obtain the final structured extraction results.

To compare with the ChatGPT, we employed the multi-turn question answering (MTQA) IE system developed in our previous work \cite{Hu2022} as the baseline. This MTQA IE system transforms the traditional named entity recognition and relation extraction tasks into a multi-turn question-answering task using BERT. It first identifies the head entities in the report during the first round of question answering and then generates the tail entity questions using the extracted results and question template. Based on the tail entity questions, the MTQA IE model extracts the answers from the CT reports during the second round of question answering. We also developed a post-processing module to convert the extracted results to a structured format.

To explore whether the instructions in the prompt can affect the extraction results, we first analyze the extraction results obtained using the prompt template in Figure \ref{figure1} and summarize some of the common errors. We then redesign the prompt template by adding some instructions based on medical knowledge as the new prompt template. 

Since the language model produces different answers to the same prompt each time, we randomly selected 100 CT reports and repeated the question-answering procedure three times for each to test the consistency of the extraction results. 

We selected the accuracy, precision, recall, and F1 score as the metrics to evaluate the IE performance.

\section{Results}

\subsection{Zero-shot IE performance}

In this study, we first analyze the zero-shot IE performance of ChatGPT for CT reports. Table \ref{table2} shows the IE performances of the baseline MTQA model and ChatGPT. One response of ChatGPT is listed in the supplement. For questions about tumor long and short diameters, tumor lobulation, pleural invasion or indentation, and mediastinal lymph node status, ChatGPT achieved competitive performances in comparison with the MTQA model even without any model training or prompting few-shot examples, which is quite surprising. But, for questions about tumor density (solid, pure ground glass, or mixed ground glass) and spiculation, ChatGPT did not show good performances compared with the MTQA model. 

We further analyzed the errors in the extraction results of ChatGPT for each question and summarized some reasons including: 

\begin{itemize}
\item For questions about tumor density (No.4, 5, 6), many errors due to that CT reports often describe multiple tumors with solid and ground glass densities, resulting in an answer of "True" to both solid and ground-glass questions (No.4 and 5) and an answer of "False" to mixed ground-glass question (No.6). Besides, some mixed ground glass tumors are described as "a solid-ground glass nodule" in the CT reports, resulting in the same errors ("True" for No.4 and 5, "False" for No.6).
\item For the question about tumor spiculation (No.7), we find that ChatGPT always sets an answer of "True" when the CT report describes that the tumor has a "streaky shadow".
\item For the question about tumor lobulation (No.8), ChatGPT will set an answer of "True" when the CT report mentions that the tumor has an "irregular shape".
\item For questions about pleural invasion or indentation (No.9), "True" will be set when the CT report mentions that there is a "pleural thickening". Besides, terms such as "horizontal fissure", "oblique fissure", and "interlobar fissure" are sometimes used to describe the interlobar pleura in the CT reports, which ChatGPT can not recognize, resulting in a "False" answer.
\item For questions about the mediastinal and hilar lymph nodes (No.10 and 11), we find that CT reports often describe "hilar lymph nodes" as "mediastinal group 10 lymph nodes". With this description, ChatGPT will mistake hilar lymph nodes for mediastinal lymph nodes and set question 10 to "True" and question 11 to "False".
\end{itemize}

For these extraction error reasons, in the following section, we will try to do the prompt engineering by adding some prior medical knowledge into part III of the base prompt template to explore whether the new prompt will reduce the extraction errors.

\begin{table}[h]
\small
\setlength{\tabcolsep}{3pt}
\caption{The IE performances of the MTQA and ChatGPT}
\label{table2}
\begin{center}
\begin{tabular}{clllllllll}
\toprule
\multirow{2}{*}{No.} & \multirow{2}{*}{Question}    & \multicolumn{4}{l}{ChatGPT}  & \multicolumn{4}{l}{MTQA}\\
 
& & Accuracy & Precision & Recall & F1 score & Accuracy & Precision & Recall & F1 score \\ 
\midrule
1  & Tumor location                              & 0.985	& 0.951	& 0.987	& 0.966 & 0.995	& 0.997	& 0.990	& 0.993 \\
2  & Tumor long diameter                         & 0.960	& 0.960	& 1.000	& 0.980 & 0.957	& 0.957 & 1.000 & 0.978 \\
3  & Tumor short diameter                        & 0.953	& 0.953	& 1.000	& 0.976 & 0.956	& 0.956	& 1.000	& 0.978 \\ 
4  & Is the tumor solid                          & 0.948	& 0.990	& 0.938	& 0.963 & 0.960	& 0.993	& 0.951	& 0.972 \\
5  & Is the tumor ground-glass opacity           & 0.894	& 0.598	& 0.873	& 0.710 & 0.960	& 0.795	& 0.984	& 0.879 \\
6  & Is the tumor mixed ground-glass opacity     & 0.924	& 0.774	& 0.591	& 0.670 & 0.946	& 0.802	& 0.773	& 0.787 \\
7  & Does the tumor have spiculations            & 0.877	& 0.726	& 0.996	& 0.840 & 0.976	& 0.988	& 0.938	& 0.963 \\ 
8  & Does the tumor have lobulations             & 0.954	& 0.860	& 1.000	& 0.925 & 0.906	& 1.000	& 0.667	& 0.800 \\
9  & Is there pleural invasion or indentation    & 0.913	& 0.899	& 0.917	& 0.908 & 0.854	& 0.802	& 0.915	& 0.855 \\
10 & Are mediastinal lymph nodes enlarged        & 0.950	& 0.929	& 0.923	& 0.926 & 0.941	& 0.961	& 0.859	& 0.907 \\ 
11 & Are hilar lymph nodes enlarged              & 0.904	& 0.775	& 0.867	& 0.819 & 0.937	& 1.000	& 0.749	& 0.856 \\
\bottomrule
\end{tabular}
\end{center}
\end{table}

\subsection{Analysis of prompts on the IE performance}

\begin{figure}[h]
\centering
\includegraphics[width=\textwidth]{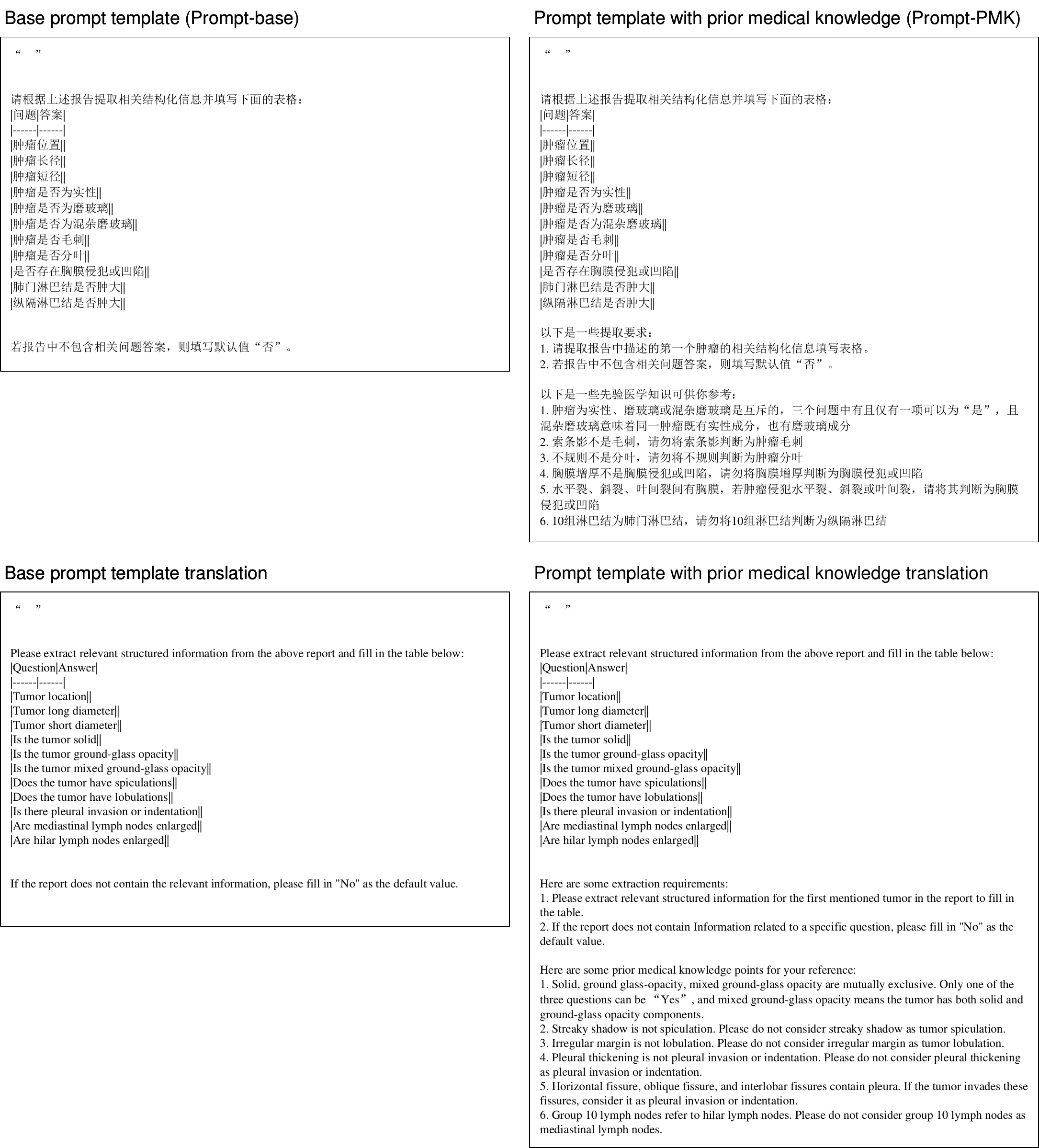}
\caption{Base prompt template and prompt template with prior medical knowledge.}
\label{figure2}
\end{figure}

In this section, we redesign the base prompt template (prompt-base) by adding prior medical knowledge to explore whether the new prompt can alleviate the wrong extraction problem. The new prompt template with prior medical knowledge (prompt-PMK) is shown in Figure \ref{figure2}. We first add instruction as a general requirement to ask ChatGPT to extract the structured information of the first tumor described in the CT report. Then, we supplement 6 instructions after the general requirements to provide ChatGPT with additional prior medical knowledge to constrain its extraction results. Using the new prompt template, we conduct the experiment to obtain the new extraction results. Table \ref{table3} shows the IE performances of ChatGPT with the prompt-base and prompt-PMK.

ChatGPT with prompt-PMK achieves significant improvements on questions about tumor spiculation (No.7), lobulation (No.8), and pleural invasion or indentation (No.9). However, for questions about tumor density (No.4, 5, 6) and lymph node status (No.10, 11), ChatGPT with prompt-PMK is not superior to ChatGPT with prompt-base. We further explore the reason for ChatGPT with prompt-PMK performance degradation on these questions. For questions about tumor density, ChatGPT does output more mutually exclusive answers, but it usually keeps the first question (No.4) True and the remaining two questions (No.5 and 6) False, which leads to a lower precision value for question No.4 and lower recall values for question No.5 and 6. For questions about lymph node status, ChatGPT is unable to fully recognize "mediastinal group 10 lymph nodes" as "hilar lymph nodes", despite our new instruction requiring ChatGPT to do so. Moreover, we also find that when mediastinal lymph nodes are described together with hilar lymph nodes, such as "mediastinal group 5, 6 and hilar lymph nodes", ChatGPT with prompt-PMK often ignores mediastinal lymph nodes, resulting in significant performance degradation in recall value of question No.10. According to the experimental results,  adding additional prior medical knowledge into the prompt may promote the information extraction for some questions, but may have a negative effect on some questions.

\begin{table}[h]
\small
\setlength{\tabcolsep}{3pt}
\caption{The IE performances of ChatGPT with different prompts}
\label{table3}
\begin{center}
\begin{tabular}{clllllllll}
\toprule
\multirow{2}{*}{No.} & \multirow{2}{*}{Question}    & \multicolumn{4}{l}{ChatGPT+Prompt-base}  & \multicolumn{4}{l}{ChatGPT+Prompt-PMK}\\
 
& & Accuracy & Precision & Recall & F1 score & Accuracy & Precision & Recall & F1 score \\ 
\midrule
1   & Tumor location                              & 0.985	& 0.951	& 0.987	& 0.966 & 0.982	& 0.945	& 0.979	& 0.959 \\
2   & Tumor long diameter                         & 0.960	& 0.960	& 1.000	& 0.980 & 0.965	& 0.965	& 1.000	& 0.982 \\
3   & Tumor short diameter                        & 0.953	& 0.953	& 1.000	& 0.976 & 0.959	& 0.959	& 1.000	& 0.979 \\ 
4   & Is the tumor solid                          & 0.948	& 0.990	& 0.938	& 0.963 & 0.885	& 0.888	& 0.962	& 0.924 \\
5   & Is the tumor ground-glass opcity            & 0.894	& 0.598	& 0.873	& 0.710 & 0.897	& 0.633	& 0.738	& 0.681 \\
6   & Is the tumor mixed ground-glass opcity      & 0.924	& 0.774	& 0.591	& 0.670 & 0.906	& 0.895	& 0.309	& 0.459 \\
7   & Does the tumor have spiculations            & 0.877	& 0.726	& 0.996	& 0.840 & 0.986	& 0.989	& 0.967	& 0.978 \\ 
8   & Does the tumor have lobulations             & 0.954	& 0.860	& 1.000	& 0.925 & 0.976	& 0.923	& 1.000	& 0.960 \\
9   & Is there pleural invasion or indentation    & 0.913	& 0.899	& 0.917	& 0.908 & 0.935	& 0.922	& 0.942	& 0.932 \\
10  & Are mediastinal lymph nodes enlarged        & 0.950	& 0.929	& 0.923	& 0.926 & 0.907	& 0.932	& 0.778	& 0.848 \\ 
11  & Are hilar lymph nodes enlarged              & 0.904	& 0.775	& 0.867	& 0.819 & 0.909	& 0.810	& 0.829	& 0.820 \\
\bottomrule
\end{tabular}
\end{center}
\end{table}

\subsection{Consistency of extraction results}

\begin{figure}[ht]
\centering
\includegraphics[width=0.8\textwidth]{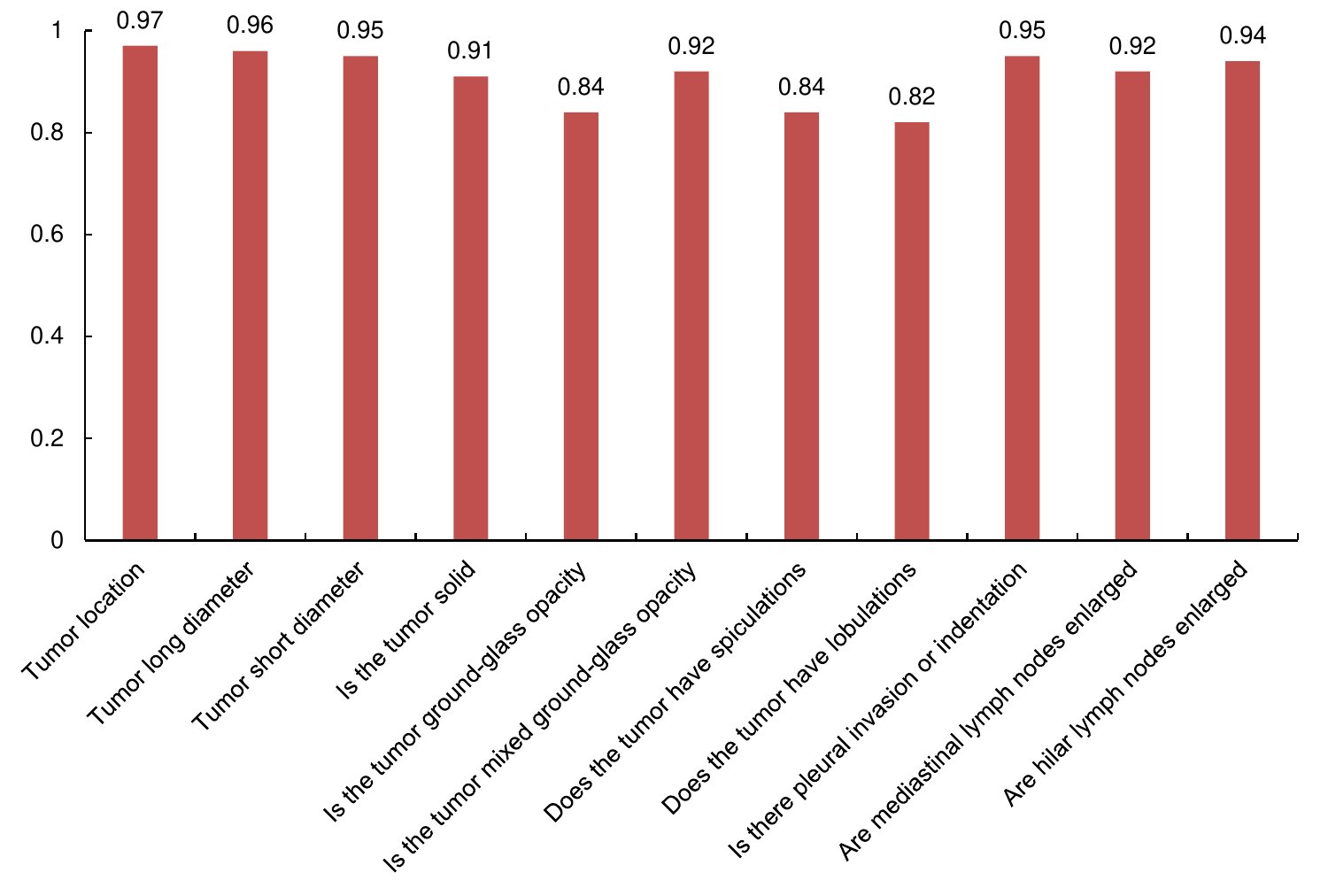}
\caption{The consistency of ChatGPT's extraction results.}
\label{figure3}
\end{figure}

Since LLMs may produce different responses for the same prompt, we conducted experiments to explore the consistency of ChatGPT's extraction results. We randomly selected 100 CT reports and used the base prompt template to generate the prompts. Then, we input each prompt three times to obtain the extraction results. We analyzed the extraction results and showed the proportion of all three results being the same in Figure \ref{figure3}. Note that the extraction results of questions about tumor density, spiculation, and lobulation have lower consistencies compared with questions about tumor location, tumor long and short diameters, pleural invasion or indentation, and lymph node status. The possible reason is that ChatGPT with prompt-base may regard some similar words in the embedding space as synonyms, sometimes not, for tumor spiculation and lobulation, and do not have a fixed set of logical rules to determine tumor density with the keywords mentioned in the CT report.

\section{Discussion}

In this study, we employed the large language model, GhatGPT, to extract the structured information from radiological reports. Experimental results indicated that ChatGPT can achieve competitive performances on several tasks compared with the baseline IE system. Using LLMs to extract information from clinical text may become a new paradigm as it does not need any corpus annotation and model fine-tuning, which is the most significant advantage over traditional IE systems. However, note that the LLMs still have some limitations.

The IE performance of ChatGPT is heavily dependent on its understanding and reasoning ability. For some simple questions such as tumor location, long and short diameter, spiculation, lobulation, and pleural invasion or indentation, ChatGPT can achieve good extraction results, even without any labeled data for guidance, which indicates that ChatGPT has a certain ability to understand the corpus and the IE questions. Most of the errors in the extraction results of the above questions are due to understanding some similar words as synonyms, such as "streaky" as "spiculation", "irregular" as "lobulation", "pleural thickening" as "pleural invasion or indentation". Or do not understand some words as synonyms, such as "horizontal fissure", "oblique fissure", and "interlobar fissure", not as "interlobar pleura". These errors can be easily avoided by adding some instructions to the prompt. For more difficult questions such as tumor density and lymph node status, ChatGPT experiences serious challenges and does not outperform the baseline IE approaches. Although we supplement additional instruction for these questions in the prompt, ChatGPT does not effectively utilize them. For example, in the prompt-PMK, we point out that the three questions about tumor density are mutually exclusive, but there are often multiple questions with "True" answers. And ChatGPT still can not extract only the information of the first described tumor in the CT report, nor can it identify "mediastinal 10 lymph nodes" and "halir lymph nodes" as the same concept. The problems indicate that ChatGPT still has limitations in understanding and reasoning some slightly complex questions.

Besides, as an LLM, the consistency of ChatGPT's outputs is also critical for the IE. One problem is that ChatGPT may produce different answers each time for the same prompt. Another problem is that ChatGPT may give answers in unanticipated free-text form, which makes it quite challenging to structure the extraction results. How to constrain the output form of the LLM should be explored in the future.

Another question is how to protect the privacy of medical data. The current ChatGPT is deployed by the openAI on their servers. If we want to use it to extract information from medical text, we need to input the original medical text into the ChatGPT, which may increase the risk of medical data leakage. One possible solution is to deploy a model like ChatGPT in the hospital's internal network. However, the computing hardware of the general hospital can not support running such a large model. Therefore, balancing the size of the model parameters and the ability of understanding and reasoning is crucial for the localization of LLMs, which deserves further study.

\section{Conclusion}

In this study, we proposed a zero-shot IE approach using ChatGPT to extract information from radiological reports. Experimental results show that ChatGPT can achieve competitive performances for some IE questions compared with the baseline IE model. We can further improve the IE performance of ChatGPT for some simple questions by adding prior medical knowledge to the prompt. But for some slightly complex questions, ChatGPT can not benefit from the added knowledge and even achieve worse IE results.

\section*{Acknowledgments}

This study was supported by the Beijing Natural Science Foundation (L222020) and the Key Research Project of Zhejiang Lab (2022PG0AC02).

\section*{Supplements}

\url{https://chat.openai.com/share/5866405b-70a4-4ef1-b854-ac1ba3f34ec0}

\url{https://chat.openai.com/share/726fe950-6a37-41c0-a01c-f0e0ec3ae973}

\end{document}